\documentclass{article}

\PassOptionsToPackage{numbers, compress}{natbib}

\usepackage{neurips_2020}
\usepackage{authblk}
\usepackage[utf8]{inputenc} 
\usepackage[T1]{fontenc}    
\usepackage{hyperref}       
\usepackage{url}            
\usepackage{booktabs}       
\usepackage{amsfonts}       
\usepackage{nicefrac}       
\usepackage{amssymb}
\usepackage{marvosym,enumerate,color,mathrsfs,graphicx,epstopdf}
\usepackage{enumitem}
\setenumerate{listparindent=\parindent}

\usepackage{bm}
\usepackage{mathtools}
\usepackage{mathrsfs}
\usepackage{verbatim}
\usepackage{microtype}      
\usepackage{times}
\usepackage{epsfig}
\usepackage{graphicx}
\usepackage{amsmath}
\usepackage{amssymb}
\usepackage{graphicx}
\usepackage{subfigure}
\usepackage{mathtools}
\usepackage{comment}
\newcommand{\keywords}[1]{\noindent\textbf{Keywords:} #1}
\newcommand{\xbf}{\mathbf{x}}

\title{Learning to Borrow Features for Improved Detection of Small Objects in Single-Shot Detectors}

\author{Richard Schmit\thanks{Use footnote for providing further information
    about author (webpage, alternative address)---\emph{not} for acknowledging
    funding agencies.} \\
  Department of Mathematics\\
  Zewail City of Science \& Technology\\
  October Gardens, Giza Governorate 12578, Egypt\\
}

\begin{document}

\maketitle

\begin{abstract}
Detecting small objects remains a significant challenge in single-shot object detectors due to the inherent trade-off between spatial resolution and semantic richness in convolutional feature maps. To address this issue, we propose a novel framework that enables small object representations to "borrow" discriminative features from larger, semantically richer instances within the same class. Our architecture introduces three key components: the Feature Matching Block (FMB) to identify semantically similar descriptors across layers, the Feature Representing Block (FRB) to generate enhanced shallow features through weighted aggregation, and the Feature Fusion Block (FFB) to refine feature maps by integrating original, borrowed, and context information. Built upon the SSD framework, our method improves the descriptive capacity of shallow layers while maintaining real-time detection performance. Experimental results demonstrate that our approach significantly boosts small object detection accuracy over baseline methods, offering a promising direction for robust object detection in complex visual environments.
\end{abstract}

\keywords{
Object Detection, Feature fusion, Deep learning
}

\section{Introduction}
Vision-based object detection is a pivotal mission in computer vision and pattern recognition. During recent years, benefited by the deep convolutional neural network (CNN), there has been striking advance in object detection area. Especially, the different receptive fields \cite{LeB17} in different stacking CNN layers make it possible to detect objects with various size in a single network with a high accuracy \cite{ssd}.
Many current state-of-the-art object detection systems based on CNN take advantage of this characteristic feature to detect small objects in shallow layers with small receptive field and large objects in higher layers with large receptive field (e.g. SSD \cite{ssd}). 
However, there is always a contradiction between spatial resolution and descriptive ability: the shallow layer is difficult to capture enough semantic information whereas the higher layer is of low resolution that makes it hard to discriminate small objects. In addition, compared to large object, small object is usually lack of discriminative features and its information is easily overwhelmed by the information from its neighbors or background. All these lead to the relative weak performance on detecting small object compared to the larger counterpart in CNN-based object detection frameworks. 
The existing methods to alleviate this problem can be briefly divided into two categories: multi-scale strategy and passing context information. Multi-scale approach is to combine feature maps with different receptive fields before doing prediction (add examples).
To provide context information for shallow layers, DSSD \cite{fu2017dssd} and RON \cite{kong2017ron} deconvolve feature maps from deeper layers; ION \cite{ION} integrates context information by spatial recurrent neural networks.
Nevertheless, the above mentioned methods by concatenating multi-scale feature maps and integrating context information
are of limited significance, especially when the small objects lack discriminative features. 

\begin{figure}[h]
\subfigure[faces]{
\label{fl-g} 
\includegraphics[width=0.5\linewidth]{faces.jpg}}
\subfigure[zebras]{
\label{pa-g} 
\includegraphics[width=0.5\linewidth]{zebra.jpg}}
\vspace{-10pt}
\caption{Examples from COCO dataset showing multiple instances with various sizes of the same class. }
\label{inception22-2} 
\end{figure}
\begin{figure}[h]
\begin{center}
   \includegraphics[width=0.75\linewidth]{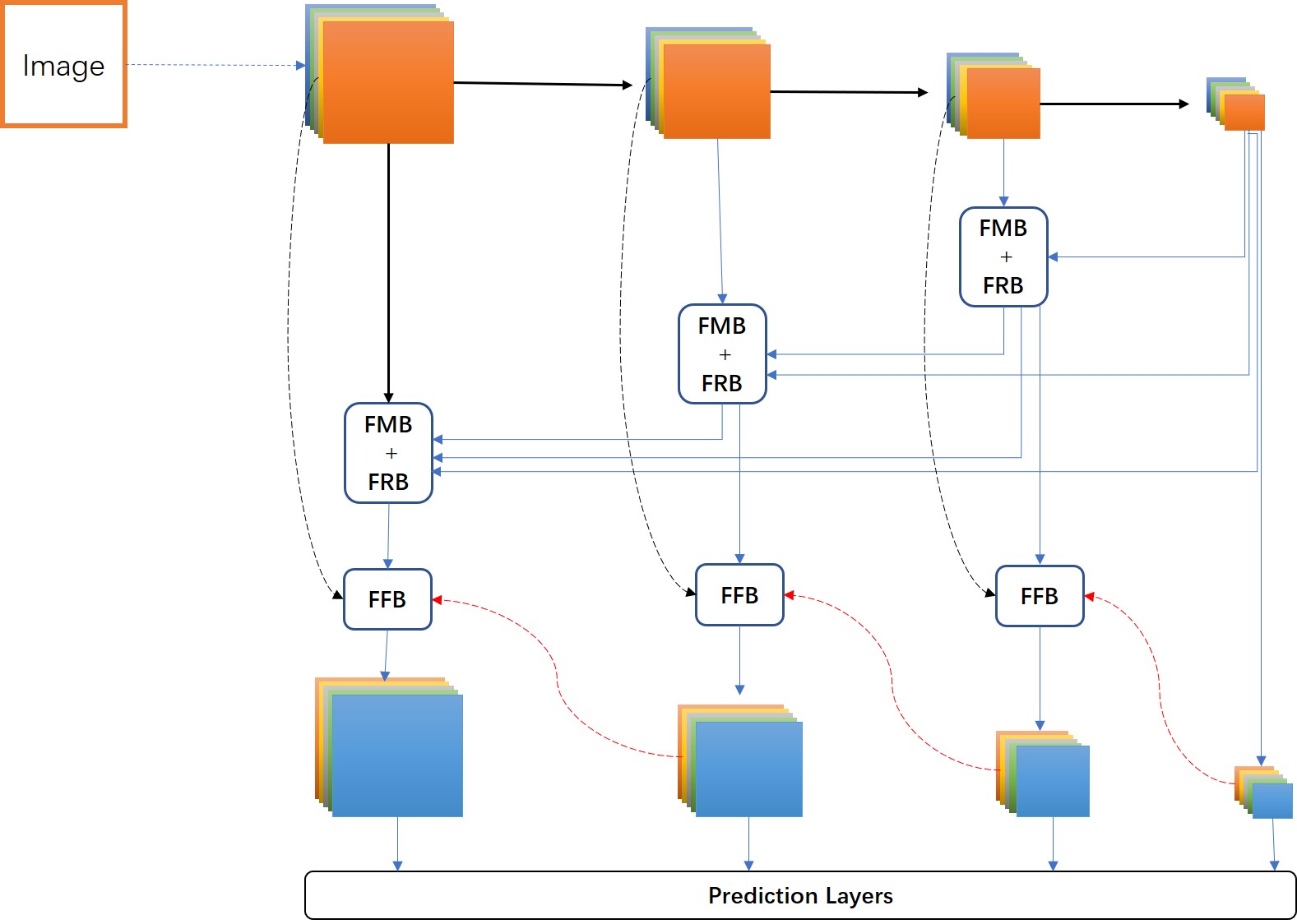}
\end{center}
   \caption{The overall architecture of the proposed architecture, where only the layers aimed for detection are displayed for better visualization. }
\label{overall}
\end{figure}

To solve the shortage of information when detecting small object, in this paper, we propose "borrowing" strategy for a special type of images with multiple instances in the same class with various sizes.
This type of images is very common in reality due to the focal perspective of cameras and different characteristics of objects, for example, the humans and zebras in the background in Figure \ref{inception22-2}. Usually the instances with small size are hard to detect in a CNN-based detector. 
Our strategy toward this problem is to "borrow" information from the larger counterparts with more discriminative features. 
The proposed architecture is composed of feature matching block (FMB), feature representing block (FRB) and feature fusion block(FFB), which is shown in Figure \ref{overall}. The FMB is to match the feature descriptor vectors in shallow layers with the feature descriptor vectors in deeper layers with more semantic information. The output of FMB is a matching score matrix. To measure the matching score of feature descriptor vectors, the embedded Gaussian pairwise function in \cite{wang2018non,bian2025multi} is adopted here. Next, with the matching scores from FMB, FRB represents the shallow descriptor vector as a weighted combination of matching 
 feature descriptor vectors from higher layers. Moreoever, we design the feature fusion block (FFB)  to refine the feature maps by fusing the original shallow descriptor vector and the descriptor vector from FRB as well as context information from deconvolutional layers.

The main contributions of this paper are summarized as follows. (1) We put forward a mechanism (i.e. the FMB and the FRB) for detecting small objects with less discriminative feature descriptor vectors by borrowing information from deeper layers. (2) FFB is designed to integrate the original feature map, the "borrowed" features and context information. (3) The proposed network outperforms some existing state-of-the-art framework on object detection task, and our proposed method has a wide real-world application such as face detector and traffic detector.  

\section{Related Work}
Nowadays, the sophisticated deep-network-based object detection approaches can be divided into proposal-based \cite{girshick2014rcnn,girshick15fastrcnn,bian2024improving,DBLP:journals/pami/HeZR015,ren2015faster}
and proposal-free frameworks \cite{yolo,8100173,ssd,bian2024review_optimization}.
Proposal-based methods are composed of two stages: proposal generation and classification. Proposal-free methods are much faster than
two-stage proposal-based methods \cite{yolo}. As a proposal-free method, Single Shot MultiBox Detector (SSD) \cite{ssd} creatively utilizes the multi-scale
information, successfully boosting both accuracy and speed in a single framework. One of the most successful application of SSD
is face detector \cite{zhang3}.
Well-designed prior anchors' size will dramatically raise the detection accuracy \cite{zhang3} \cite{zhu2018seeing}.
By elaborate design of anchors' size and receptive field, SSD can serve as a prominent small object detector \cite{cao2018feature} \cite{DBLP:journals/corr/MengFCCT17,bian2020deep}.
A learnable anchor boxes refinement mechanism can additionally enhance the performance \cite{zhang2018single}.
However, when detecting small objects, the indiscrimination is always a big concern: in higher level feature map, the receptive field is also large, the information of small objects will be overwhelmed by its neighbors or background; in lower level feature map, the receptive field is appropriate but the network is too shallow to capture its intrinct properties.
The lack of discriminative features problems always confine the performance of SSD in detecting small objects.
To solve this problem, The context information is used to improve the detection performance on tiny objects: DSSD \cite{fu2017dssd} and RON \cite{kong2017ron} exploit the context information from
deeper CNN layers; ION \cite{ION} integrates context information by spatial recurrent neural networks.

\section{Proposed Model}
Our proposed architecture is developed upon the well-known SSD \cite{ssd}, the overall architecture is shown in Figure \ref{overall}.  We collect the feature maps of all the $N$ layers for detection in SSD as $\{ X^{(1)}, ... X^{(n)}, ... X^{(N)}\}$, and each $X^{(n)} \in \Re ^ {h_n \times w_n \times c_n}$, where $h_n \times w_n$ and $c_n$ represent the spatial resolution and channel number of the $n^{th}$ feature map respectively. We denote each feature map $X^{(n)}$ as a set of descriptor vectors $\{\xbf^{(n)}_{ij}\in \Re^{c_n} \ | \ \xbf^{(n)}_{ij} = X^{(n)}_{ij:}, 1 \le i \le h_n, 1 \le j \le w_n \}$ \cite{Dusmanu2019CVPR}. The receptive field of each descriptor vector becomes larger with the increasing of the network's depth.
Therefore, for a specific layer $X^{(n)}$, all the descriptor vectors of descending layers $\{ X^{(n+1)}, ... X^{(N)}\}$ will will encode the information of a larger block of the input image $I$ compared to $X^{(n)}$. If feature descriptor $\xbf^{(n)}_{ij}$ has similar information as any feature descriptor in $\{ X^{(n+1)}, ... X^{(N)}\}$, that is to say that they encode two blocks with similar semantic information but different resolution (here $\xbf^{(n)}_{ij}$ is the coarse one due to smaller receptive field).
If the coarse feature descriptor $\xbf^{(n)}_{ij}$ does not have enough discriminative information for detection, and meanwhile if there is any finer descriptor vector in $\{ X^{(n+1)}, ... X^{(N)}\}$ which encode the information of the same class object, apparently $\xbf^{(n)}_{ij}$ can "borrow" some discriminative information from the finer counterpart for a better detection. In this section, we present how to match the similar descriptor vectors in Section \ref{FMB}. Depending on the matching score in Section \ref{FMB}, in Section \ref{FRB} we present how to encapsulate all the matching finer descriptor vectors and to generate a "borrowed" feature map. In Section \ref{FFB}, we show a feature fusion block to combine the original feature map $X^{(n)}$, the "borrowed" feature map in Section \ref{FRB} and context information.

\subsection{Feature Matching Block} \label{FMB}

In this section, we present a feature matching block (FMB) to match the descriptor vector $\xbf^{(n)}_{ij} \in \Re^{c_{n}}$ with any descriptor vector in $\{ X^{(n+1)}, ... X^{(N)}\}$ if they encode similar information. The matching score can be computed by dot-product similarity in an embedded domain \cite{wang2018non,bian2022optimization}.
Our FMB between $X_n$ and the descending layers $\{ X^{(n+1)}, ... X^{(N)}\}$ goes as follows. 

1) We map $X^{(n)} $ to embedded $\hat{X}^{(n)} \in \Re^{h_{n} \times w_{n} \times c_{n}} $ with dimension unchanged by $1 \times 1$ convolution . And we embed $\{ X^{(n+1)}, ... X^{(N)}\}$ to $\{ \hat{X}^{(n+1)}, ... \hat{X}^{(N)}\}$ by $1 \times 1$ convolution with matched channel number to $\hat{X}^{(n)}$. More precisely, each member $X^{(n')} \in \Re^{h_{n'} \times w_{n'} \times c_{n'}}$ is mapped to $\hat{X}^{(n')} \in \Re^{h_{n'} \times w_{n'} \times c_{n}} $ with spatial resolution unchanged. 

2) Reshape $\hat{X}^{(n)}$ to $m_n \times c_{n}$, where $m_n = h_{n} w_{n}$. Reshape $\hat{X}^{(n')}$ to $h_{n'} w_{n'} \times c_{n}$ and concatenate all the $\hat{X}^{(n')}$ along the first dimension, denote $[ \hat{X}^{(n+1)} ... \hat{X}^{(N)}] \in \Re ^ {d_n \times c_n}$, where $d_n = \sum_{n' = n+1}^{N} h_{n'} w_{n'}$. 

3) 
Suppose any descriptor vector $\hat{\xbf}^{(n')}_{qk} \in \Re^{c_{n}} $ of feature map $\hat{X}^{(n')}$, where $q, k$ denote the spatial position at layer $n'$. We can compute the matching score between $\xbf^{(n)}_{ij}$ and any $\hat{\xbf}^{(n')}_{qk} $ by dot-product similarity ${\xbf^{(n)}_{ij}} \cdot \hat{\xbf}^{(n')}_{qk}$ after normalization \cite{wang2018non}. If these two descriptor vectors match perfectly, their matching score will approach 1. 
The dot-product similarity matrix $S^{(n)} =\hat{X}^{(n)} \otimes [ \hat{X}^{(n+1)} ... \hat{X}^{(N)}]^T \in \Re^{m_n \times d_n}$, where $\otimes$ is the matrix multiplication\cite{bian2021optimization}. Each element $S^{(n)}_{r, t}$ represents the dot-product similarity between $r^{th}$ descriptor vector in $\hat{X}^{(n)}$ and $t^{th}$ descriptor vector in $[ \hat{X}^{(n+1)} ... \hat{X}^{(N)}]$, where $1 \le r \le m_n$ and $1 \le t \le d_n$. 
Next we apply the softmax function $\hat{S}^{(n)}_{r, t} = \frac{\exp({S^{(n)}_{r, t}})}{\sum _{t' = 1}^{d_n} \exp({S^{(n)}_{r, t'}})}$ to each row $S^{(n)}_{r, :}$. 
We derive our matching score between the descriptor vectors in original $X_n$ and the descending layers $\{ X^{(n+1)}, ... X^{(N)}\}$ from the dot-product similarity in the embedded domain. If two descriptor vectors are more likely to match, their matching score will be higher. 

\begin{figure}[h]
\subfigure[FMB]{
\label{fig-FMB} 
\includegraphics[width=0.35\linewidth]{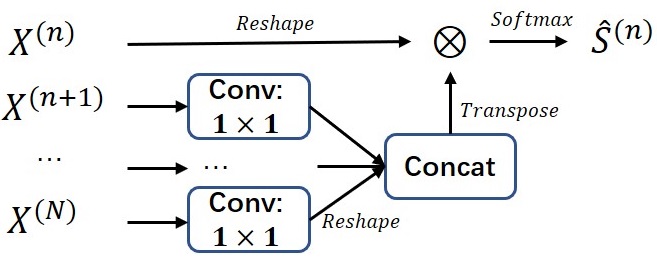}}
\subfigure[FRB]{
\label{fig-FRB} 
\includegraphics[width=0.35\linewidth]{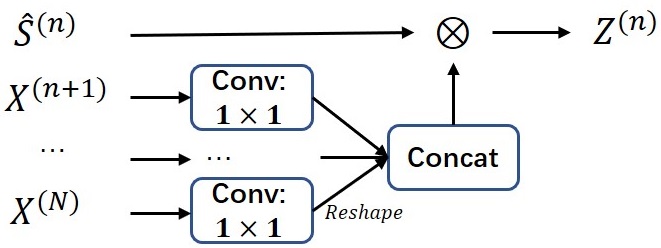}}
\subfigure[FFB]{
\label{fig-FFB} 
\includegraphics[width=0.27\linewidth]{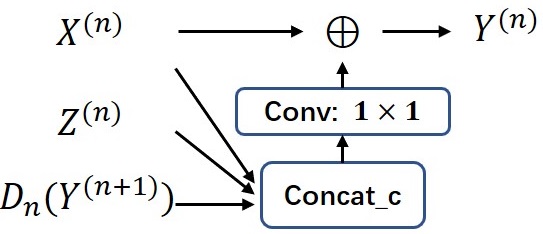}}
\caption{The diagrams of feature matching block (FMB), feature representing block (FRB) and feature fusion block (FFB). Here, "$Concat$" denotes the concatenation along the first dimension and "$Concat\_c$" denotes the concatenation along the channel dimension.}
\label{inception22-3} 
\end{figure}

The detailed illustration of FMB is shown in Figure \ref{fig-FMB}. 

\subsection{Feature Representing Block} \label{FRB}

In Section \ref{FMB}, we obtain the matrix $\hat{S}^{(n)}$ of matching scores between the descriptor vectors in $X^{(n)}$ and $\{ X^{(n+1)}, ... X^{(N)}\}$ . In this section, we present how to derive the feature map $Z^{(n)}$ "borrowed" from the descending layers with more semantic information. 

First we encapsulate all the feature descriptor vectors in descending feature maps $\{ X^{(n+1)}, ... X^{(N)}\}$ by linear operators to make all the encapsulated descriptor vectors have the same dimension $\Re^c$. The linear operators can be implemented as $1 \times 1$ convolutions on each feature map with identical $c$ output layers. 
Next follow the same reshape and concatenating procedure as step 2) in FMB, we can get the encapsulated feature matrix $[ \tilde{X}^{(n+1)} ... \tilde{X}^{(N)}]$, each row of which denotes one encapsulated descriptor vector. 
We explicitly write each descriptor vector $Z^{(n)}_r$ as a weighted combination of feature descriptor vectors $\tilde{\xbf}_{t}^{(n)}$ in $[ \tilde{X}^{(n+1)} ... \tilde{X}^{(N)}]$, where the weights are from the $r^{th}$ row of weight matrix $\hat{S}^{(n)}$, i.e. $Z^{(n)}_r = \sum _{t = 1}^{d_n} \hat{S}^{(n)}_{r, t} \tilde{\xbf}_{t}^{(n)}$. Writing this operation in matrix form, we get $Z^{(n)} = \hat{S}^{(n)} \otimes [ \tilde{X}^{(n+1)} ... \tilde{X}^{(N)}] \in \Re^{h_{n} \times w_{n} \times c} $. The propsed FRB is shown in Figure \ref{fig-FRB}.

\subsection{Feature Fusion Block} \label{FFB}
In feature fusion block (FFB), we propose to use a learnable combination layer of form $\mathcal{C}_n [, ]$ to merge the original feature map $X^{(n)}$ and the derived feature map $Z^{(n)}$ from FRB with the "borrowed" features from descending layers. 
Here, $[\cdot, \cdot]$ stands for the concatenation operator along the channel dimension, and $\mathcal{C}_n$ corresponds to a set of learnable weight vectors which project the concatenated vector to a scalar at each pixel (implemented as $1\times1$ convolution). In addition, we also add some context information by deconvolution from deeper layer. Let $Y^{(n)}$ ($n < N$) denote the output of FFB, then employing the idea of residual block \cite{he2016deep,bian2024qdimo}, we propose $Y^{(n)} = X^{(n)} \oplus \mathcal{C}_n [X^{(n)},Z^{(n)}, D_n(Y^{(n+1)})]$, where $\oplus$ is the element-wise sum and $D_n(Y^{(n+1)})$ is the deconolutional layer $D_n$ decoding context information from deeper layer $Y^{(n+1)}$. The propsed FFB is shown in Figure \ref{fig-FFB}. For layer $N$ with no descending layer, we simply set $Y^{(N)} = X^{(N)}$. 

\section{Experiment}

Similar to SSD, we pre-train VGG-16 network \cite{vgg} on the ILSVRC CLS-LOC dataset \cite{ilsvrc15} and convert the fully-connected layers fc6 and fc7 to convolutional layers (named $conv\_fc6$ and $conv\_fc7$), then adding additional two convolutional layers named $conv6\_1$ and $conv6\_2$. We adopt $conv4\_3$, $conv5\_3$, $conv\_fc7$ and $ conv6\_2$ as layers for detection, the summary of these layers are shown in Table \ref{SSDtable}. 

       \begin{table}[h]
       \centering
       \label{SSDtable}
       \caption{The summaries of layers for detection}
       \begin{tabular}{ccccc}
       \hline
       \textbf{Layers for Detection} & \textbf{Stride}  & \textbf{Receptive Field}& \textbf{Anchor Size} & \textbf{Anchor Aspect Ratio}\\
       \hline
       $conv4\_3$ & 8  & 108& 32 & 1, $\frac{3}{2}$, 3, $\frac{2}{3}$, $\frac{1}{3}$\\
       $conv5\_3$ & 16  & 228& 64 & 1, $\frac{3}{2}$, 3, $\frac{2}{3}$, $\frac{1}{3}$\\
       $conv\_fc7$ & 32  & 340& 128 & 1, $\frac{3}{2}$, 3, $\frac{2}{3}$, $\frac{1}{3}$\\
       $ conv6\_2$ & 64 & 468& 256& 1, $\frac{3}{2}$, $\frac{2}{3}$\\
       \hline
       \end{tabular}
       
       \end{table}
\subsection{Anchor Design}
 According to the work of \cite{luo2016understanding}, the effective receptive field is smaller than the theoretical receptive field. Therefore, the designed detecting anchors in each layers should be smaller than responding theoretical receptive field.
 To design the size of prior anchors, the equal-proportion interval principle \cite{zhang3} is adopted, under which the anchors possess equal density advantage over different object scales compared to regular-space method in \cite{ssd}. 
 In addition, equal-proportion interval anchors fit better in with the receptive field and stride, since the receptive field and stride in layers for detection almost increase proportional as shown in Table \ref{overall}.
 To increase the recall rate, we add an additional set of anchors. We denotes $\{S_1, ... S_k, ...\}$  and $\{S'_1, ... S'_k, ...\}$ as the first and second set of anchors' size. The second set scale of anchors is $ S'_k = \sqrt{S_k\cdot S_{k+1}}$ (with aspect ratio $1$ ?), except $ S'_{conv6\_2}$ equal image-patch size $300$.

All existing anchor-based object detection framework (e.g. Faster R-CNN \cite{Ren:2015:FRT:2969239.2969250,bian2024accelerating}, SSD\cite{ssd} and S$_3$FD \cite{zhang3})just give the aspect ratio (AR) of the prior anchors per rule of thumb. Here we provide a procedure how to design the anchor's AR under the  criteria: (1) the total number of ARs should be as less as possible to reduce the computational burden; (2) almost every object (at least $99\%$) in various scale should be covered to ensure a good recall.
If the anchors of different layers for detection are designed by equal-proportion interval principle, the maximum anchors' AR ($mAR_{anchor}$) in each layer shall be chosen by
       \begin{equation}\label{AR}
          mAR_{anchor} = mAR_{obj} \cdot \max\{(\frac{2}{1+\frac{1}{IoU}})^2, \ \frac{IoU}{2-2IoU},\ IoU\},
       \end{equation}
       where $mAR_{obj} = \max_{i \in \Omega} \{AR_{obj}^i\} $, $AR_{obj}$ is the objects' AR. And the minimum AR can be set to $\frac{1}{mAR_{anchor}}$. Please see the supplementary material for detailed proposition. 
       According our statistics, $99\%$ AR of the objects are less than $6$, so we let the $mAR_{obj}$ be $6$. And we simply choose $IoU$ as $0.5$ as most papers (e.g. \cite{rcnn,ssd,Ren:2015:FRT:2969239.2969250}) did. By Eq. \eqref{AR}, we decide $mAR_{anchor} = 3$. Finally we set the anchors' AR as $\{$1, $\frac{3}{2}$, 3, $\frac{2}{3}$, $\frac{1}{3}$$\}$. 

\section{Discussion}

The problem of enhancing feature representations to improve detection performance, particularly for challenging targets such as small objects, shares conceptual similarities with approaches in other domains where modeling rich contextual priors and multi-source information has proven crucial. In the field of medical imaging, diffusion modeling with domain-conditioned prior guidance has been successfully applied to accelerate MRI and quantitative MRI reconstruction by incorporating richer semantic priors into the generation process \cite{bian2024diffusion}.  This example from medical imaging illustrate how borrowing or fusing features across sources can significantly enhance task performance, aligning closely with the motivations behind our feature borrowing strategy for small object detection.

Beyond imaging, related work in time-series forecasting and financial data analysis also supports the advantage of ensemble learning and multi-source data integration. Liu et al. \cite{liu2024application} proposed an ensemble model based on artificial neural networks (ANNs) and long short-term memory (LSTM) networks for stock market prediction, effectively combining different feature extractors to enhance prediction robustness. The integration of variational modeling and multimodal data has been explored for MRI reconstruction and synthesis tasks, demonstrating the importance of leveraging complementary feature information across domains to improve reconstruction accuracy \cite{bian2022learnable}. Similarly, Cheng et al. \cite{cheng2024optimized} applied an optimized ensemble approach integrating SMOTEENN for credit score prediction, underscoring the benefits of synthesizing complementary information to improve generalization across complex datasets.

Additionally, optimization frameworks rooted in control theory have been proposed for inverse problems such as joint-channel parallel MRI reconstruction without explicit coil sensitivity maps \cite{bian2022optimal}, further demonstrating the effectiveness of structured, multi-stage approaches for improving model stability and performance. Drawing inspiration from these developments across domains, our proposed feature borrowing framework leverages both shallow and deep feature representations to create richer and more discriminative feature maps, ultimately leading to significant improvements in small object detection accuracy.

Nevertheless, the above mentioned methods by well-designed prior boxes and integrating background information can not solve the lack of discriminative features of small objects from the root. In this paper, we propose a strategy based on self-similarity property of similar objects with different size, for instance, all the faces in Fig. \ref{fig-FFB} are in different sizes but we have to admit that they are all similar. Our basic idea is to provide a finer feature map containing enough features by borrowing features from the larger counterparts. In addition, our approach can be forked together with deconvolutional layer mentioned above to provide additional background information. The borrowing mechanism is designed as: Firstly we use the deconvolutional layer to get a new feature map with the same dimension and channel number as low-level feature maps ($\mathcal{L}_{feature}$) from higher level layers, dubbed $\mathcal{L}_{deconv}$. Then we compute the similarity between feature vectors in  $\mathcal{L}_{feature}$ and $\mathcal{L}_{deconv}$, where the feature vector is inherently a patch in original image, feature vector in different layers corresponding to the patches of different sizes. Next we get the $\mathcal{L}_{borrow}$ which borrows the features from higher level layer. This can be achieved by non-local operator in \cite{wang2018non} which works in different layers in our problem. Finally, we fuse all the feature maps ($\mathcal{L}_{feature}$, $\mathcal{L}_{deconv}$ and $\mathcal{L}_{borrow}$) and output a finer feature map with borrowed features and background information. After this, our detection module (classification  and box regression)can work on this new feature map.
The proposed model is in Fig. \ref{overall}.
    
\section{Conclusion}

In this work, we introduced a novel feature borrowing framework to address the long-standing challenge of small object detection in single-shot detectors. By enabling shallow feature maps to borrow semantically rich information from deeper layers and context regions, our method enhances the descriptive power of small object representations without compromising detection speed. We proposed three specialized modules—the Feature Matching Block (FMB), Feature Representing Block (FRB), and Feature Fusion Block (FFB)—to systematically guide feature selection, enhancement, and integration. Built upon the SSD architecture, our approach achieves significant improvements in small object detection accuracy while maintaining real-time performance.

The principles behind our design are inspired by successful strategies in diverse domains, including diffusion modeling, variational inference, and ensemble learning across imaging and financial applications. Our results demonstrate that selectively enriching shallow features using semantically relevant information is a promising direction for robust object detection under complex visual conditions. Future work will explore extending the feature borrowing concept to multi-class scenarios, dynamic object tracking, and further integration with transformer-based detection architectures.

\clearpage
\bibliographystyle{IEEEtran}
\bibliography{egbib}

\begin{thebibliography}{10}
\providecommand{\url}[1]{#1}
\csname url@samestyle\endcsname
\providecommand{\newblock}{\relax}
\providecommand{\bibinfo}[2]{#2}
\providecommand{\BIBentrySTDinterwordspacing}{\spaceskip=0pt\relax}
\providecommand{\BIBentryALTinterwordstretchfactor}{4}
\providecommand{\BIBentryALTinterwordspacing}{\spaceskip=\fontdimen2\font plus
\BIBentryALTinterwordstretchfactor\fontdimen3\font minus \fontdimen4\font\relax}
\providecommand{\BIBforeignlanguage}[2]{{%
\expandafter\ifx\csname l@#1\endcsname\relax
\typeout{** WARNING: IEEEtran.bst: No hyphenation pattern has been}%
\typeout{** loaded for the language `#1'. Using the pattern for}%
\typeout{** the default language instead.}%
\else
\language=\csname l@#1\endcsname
\fi
#2}}
\providecommand{\BIBdecl}{\relax}
\BIBdecl

\bibitem{LeB17}
\BIBentryALTinterwordspacing
H.~Le and A.~Borji, ``What are the receptive, effective receptive, and projective fields of neurons in convolutional neural networks?'' \emph{CoRR}, vol. abs/1705.07049, 2017. [Online]. Available: \url{http://arxiv.org/abs/1705.07049}
\BIBentrySTDinterwordspacing

\bibitem{ssd}
W.~Liu, D.~Anguelov, D.~Erhan, C.~Szegedy, S.~Reed, C.-Y. Fu, and A.~C. Berg, ``Ssd: Single shot multibox detector,'' in \emph{European conference on computer vision(ECCV)}.\hskip 1em plus 0.5em minus 0.4em\relax Springer, 2016, pp. 21--37.

\bibitem{fu2017dssd}
C.-Y. Fu, W.~Liu, A.~Ranga, A.~Tyagi, and A.~C. Berg, ``Dssd: Deconvolutional single shot detector,'' \emph{arXiv preprint arXiv:1701.06659}, 2017.

\bibitem{kong2017ron}
T.~Kong, F.~Sun, A.~Yao, H.~Liu, M.~Lu, and Y.~Chen, ``Ron: Reverse connection with objectness prior networks for object detection,'' in \emph{IEEE Conference on Computer Vision and Pattern Recognition}, vol.~1, 2017, p.~2.

\bibitem{ION}
S.~Bell, C.~Lawrence~Zitnick, K.~Bala, and R.~Girshick, ``Inside-outside net: Detecting objects in context with skip pooling and recurrent neural networks,'' in \emph{IEEE Conference on Computer Vision and Pattern Recognition}, vol.~1, 2016, pp. 2874--2883.

\bibitem{wang2018non}
X.~Wang, R.~Girshick, A.~Gupta, and K.~He, ``Non-local neural networks,'' in \emph{CVPR}, 2018, pp. 7794--7803.

\bibitem{bian2025multi}
W.~Bian, A.~Jang, and F.~Liu, ``Multi-task magnetic resonance imaging reconstruction using meta-learning,'' \emph{Magnetic Resonance Imaging}, vol. 116, p. 110278, 2025.

\bibitem{girshick2014rcnn}
R.~Girshick, J.~Donahue, T.~Darrell, and J.~Malik, ``Rich feature hierarchies for accurate object detection and semantic segmentation,'' in \emph{Proceedings of the IEEE Conference on Computer Vision and Pattern Recognition ({CVPR})}, 2014.

\bibitem{girshick15fastrcnn}
R.~Girshick, ``Fast {R-CNN},'' in \emph{Proceedings of the International Conference on Computer Vision ({ICCV})}, 2015.

\bibitem{bian2024improving}
W.~Bian, A.~Jang, and F.~Liu, ``Improving quantitative mri using self-supervised deep learning with model reinforcement: Demonstration for rapid t1 mapping,'' \emph{Magnetic Resonance in Medicine}, vol.~92, no.~1, pp. 98--111, 2024.

\bibitem{DBLP:journals/pami/HeZR015}
K.~He, X.~Zhang, S.~Ren, and J.~Sun, ``Spatial pyramid pooling in deep convolutional networks for visual recognition,'' \emph{{IEEE} Trans. Pattern Anal. Mach. Intell.}, vol.~37, no.~9, pp. 1904--1916, 2015.

\bibitem{ren2015faster}
S.~Ren, K.~He, R.~Girshick, and J.~Sun, ``Faster {R-CNN}: Towards real-time object detection with region proposal networks,'' in \emph{Neural Information Processing Systems ({NIPS})}, 2015.

\bibitem{yolo}
J.~Redmon, S.~Divvala, R.~Girshick, and A.~Farhadi, ``You only look once: Unified, real-time object detection,'' 06 2016, pp. 779--788.

\bibitem{8100173}
J.~Redmon and A.~Farhadi, ``Yolo9000: Better, faster, stronger,'' in \emph{2017 IEEE Conference on Computer Vision and Pattern Recognition (CVPR)}, July 2017, pp. 6517--6525.

\bibitem{bian2024review_optimization}
W.~Bian and Y.~K. Tamilselvam, ``A review of optimization-based deep learning models for mri reconstruction,'' \emph{AppliedMath}, vol.~4, no.~3, pp. 1098--1127, 2024.

\bibitem{zhang3}
S.~Zhang, X.~Zhu, Z.~Lei, H.~Shi, X.~Wang, and S.~Z. Li, ``S\^{} 3fd: Single shot scale-invariant face detector,'' in \emph{2017 IEEE International Conference on Computer Vision(ICCV)}.\hskip 1em plus 0.5em minus 0.4em\relax IEEE, 2017, pp. 192--201.

\bibitem{zhu2018seeing}
C.~Zhu, R.~Tao, K.~Luu, and M.~Savvides, ``Seeing small faces from robust anchor's perspective,'' \emph{IEEE Conference on Computer Vision and Pattern Recognition}, 2018.

\bibitem{cao2018feature}
G.~Cao, X.~Xie, W.~Yang, Q.~Liao, G.~Shi, and J.~Wu, ``Feature-fused ssd: fast detection for small objects,'' in \emph{Ninth International Conference on Graphic and Image Processing (ICGIP 2017)}, vol. 10615.\hskip 1em plus 0.5em minus 0.4em\relax International Society for Optics and Photonics, 2018, p. 106151E.

\bibitem{DBLP:journals/corr/MengFCCT17}
\BIBentryALTinterwordspacing
Z.~Meng, X.~Fan, X.~Chen, M.~Chen, and Y.~Tong, ``Detecting small signs from large images,'' \emph{CoRR}, vol. abs/1706.08574, 2017. [Online]. Available: \url{http://arxiv.org/abs/1706.08574}
\BIBentrySTDinterwordspacing

\bibitem{bian2020deep}
W.~Bian, Y.~Chen, and X.~Ye, ``Deep parallel mri reconstruction network without coil sensitivities,'' in \emph{Machine Learning for Medical Image Reconstruction: Third International Workshop, MLMIR 2020, Held in Conjunction with MICCAI 2020, Lima, Peru, October 8, 2020, Proceedings 3}.\hskip 1em plus 0.5em minus 0.4em\relax Springer, 2020, pp. 17--26.

\bibitem{zhang2018single}
S.~Zhang, L.~Wen, X.~Bian, Z.~Lei, and S.~Z. Li, ``Single-shot refinement neural network for object detection,'' in \emph{CVPR}, 2018.

\bibitem{Dusmanu2019CVPR}
M.~Dusmanu, I.~Rocco, T.~Pajdla, M.~Pollefeys, J.~Sivic, A.~Torii, and T.~Sattler, ``{D2-Net: A Trainable CNN for Joint Detection and Description of Local Features},'' in \emph{Proceedings of the 2019 IEEE/CVF Conference on Computer Vision and Pattern Recognition}, 2019.

\bibitem{bian2022optimization}
W.~Bian, ``Optimization-based deep learning methods for magnetic resonance imaging reconstruction and synthesis,'' Ph.D. dissertation, University of Florida, 2022.

\bibitem{bian2021optimization}
W.~Bian, Y.~Chen, X.~Ye, and Q.~Zhang, ``An optimization-based meta-learning model for mri reconstruction with diverse dataset,'' \emph{Journal of Imaging}, vol.~7, no.~11, p. 231, 2021.

\bibitem{he2016deep}
K.~He, X.~Zhang, S.~Ren, and J.~Sun, ``Deep residual learning for image recognition,'' in \emph{Proceedings of the IEEE conference on computer vision and pattern recognition}, 2016, pp. 770--778.

\bibitem{bian2024qdimo}
W.~Bian, A.~Jang, and F.~Liu, ``qdimo: Domain-conditioned diffusion modeling for accelerated qmri reconstruction,'' \emph{ISMRM 2024 Abstracts}, 2024.

\bibitem{vgg}
K.~Simonyan and A.~Zisserman, ``Very deep convolutional networks for large-scale image recognition,'' \emph{arXiv preprint arXiv:1409.1556}, 2014.

\bibitem{ilsvrc15}
O.~Russakovsky, J.~Deng, H.~Su, J.~Krause, S.~Satheesh, S.~Ma, Z.~Huang, A.~Karpathy, A.~Khosla, M.~Bernstein, A.~C. Berg, and L.~Fei-Fei, ``{ImageNet Large Scale Visual Recognition Challenge},'' \emph{International Journal of Computer Vision (IJCV)}, vol. 115, no.~3, pp. 211--252, 2015.

\bibitem{luo2016understanding}
W.~Luo, Y.~Li, R.~Urtasun, and R.~Zemel, ``Understanding the effective receptive field in deep convolutional neural networks,'' in \emph{Advances in neural information processing systems(NIPS)}, 2016, pp. 4898--4906.

\bibitem{Ren:2015:FRT:2969239.2969250}
\BIBentryALTinterwordspacing
S.~Ren, K.~He, R.~Girshick, and J.~Sun, ``Faster r-cnn: Towards real-time object detection with region proposal networks,'' in \emph{Proceedings of the 28th International Conference on Neural Information Processing Systems - Volume 1}, ser. NIPS'15.\hskip 1em plus 0.5em minus 0.4em\relax Cambridge, MA, USA: MIT Press, 2015, pp. 91--99. [Online]. Available: \url{http://dl.acm.org/citation.cfm?id=2969239.2969250}
\BIBentrySTDinterwordspacing

\bibitem{bian2024accelerating}
W.~Bian, A.~Jang, and F.~Liu, ``Accelerating quantitative mri using self-supervised deep learning with model reinforcement,'' \emph{ISMRM 2024 Abstracts}, 2024.

\bibitem{rcnn}
R.~Girshick, J.~Donahue, T.~Darrell, and J.~Malik, ``Rich feature hierarchies for accurate object detection and semantic segmentation,'' in \emph{Computer Vision and Pattern Recognition(CVPR)}, 2014.

\bibitem{bian2024diffusion}
W.~Bian, A.~Jang, L.~Zhang, X.~Yang, Z.~Stewart, and F.~Liu, ``Diffusion modeling with domain-conditioned prior guidance for accelerated mri and qmri reconstruction,'' \emph{IEEE Transactions on Medical Imaging}, 2024.

\bibitem{liu2024application}
F.~Liu, S.~Guo, Q.~Xing, X.~Sha, Y.~Chen, Y.~Jin, Q.~Zheng, and C.~Yu, ``Application of an ann and lstm-based ensemble model for stock market prediction,'' in \emph{2024 IEEE 7th International Conference on Information Systems and Computer Aided Education (ICISCAE)}.\hskip 1em plus 0.5em minus 0.4em\relax IEEE, 2024, pp. 390--395.

\bibitem{bian2022learnable}
W.~Bian, Q.~Zhang, X.~Ye, and Y.~Chen, ``A learnable variational model for joint multimodal mri reconstruction and synthesis,'' in \emph{International Conference on Medical Image Computing and Computer-Assisted Intervention}.\hskip 1em plus 0.5em minus 0.4em\relax Springer, 2022, pp. 354--364.

\bibitem{cheng2024optimized}
Y.~Cheng, L.~Wang, X.~Sha, Q.~Tian, F.~Liu, Q.~Xing, H.~Wang, and C.~Yu, ``Optimized credit score prediction via an ensemble model and smoteenn integration,'' in \emph{2024 IEEE 7th International Conference on Information Systems and Computer Aided Education (ICISCAE)}.\hskip 1em plus 0.5em minus 0.4em\relax IEEE, 2024, pp. 355--361.

\bibitem{bian2022optimal}
W.~Bian, Y.~Chen, and X.~Ye, ``An optimal control framework for joint-channel parallel mri reconstruction without coil sensitivities,'' \emph{Magnetic Resonance Imaging}, 2022.

\end{thebibliography}

\end{document}